# Control and Monitoring System for Modular Wireless Robot

I. Firmansyah, B. Hermanto and L.T. Handoko
Group for Theoretical and Computational Physic, Research Center for Physics, Indonesian Institute of Sciences, Kompleks
Puspiptek Serpong, Tangerang 15310, Indonesia
firmansyah@teori.fisika.lipi.go.id

**Abstract**
*We introduce our concept on the modular wireless robot consisting of three main modules : main unit, data acquisition and data processing modules. We have developed a generic prototype with an integrated control and monitoring system to enhance its flexibility, and to enable simple operation through a web-based interface accessible wirelessly. In present paper, we focus on the microcontroller based hardware to enable data acquisition and remote mechanical control.*

## 1. Introduction

Automation systems have been becoming the main issue since many decades ago to improve the efficiency and to avoid unnecessary defects, especially in modern manufactures. It also plays an important role in some manufacturing processes which require high precision. On the other hand, automation systems with embedded remote robots are getting common in non-manufacturing industries where safety is the main issue, as mine explorations, etc.

Nowadays, following the advances on internet technology some groups have developed new kind of robots which can be teleoperated over web as MAX project [1] and also its successor WAX [2,3]. Either MAX or WAX are the microcontroller-based robots equipped with onboard computer, camera and microphone with main purpose to simulate telepresence. Because originally these robots were intended for handicapped persons. So the main issue is how to recognize the captured images or sounds and interpret them to be useful information for users.

We follow similar approach in developing our LIPI Wireless Robot (LWR), but with very different purposes [4]. We develop a wirelessly teleoperated robot for more serious tasks requiring telepresence for the reason of safety. For example : retrieving data directly from nuclear reactors, observations in volcanoes and so on. The LWR is unique in the sense of its ability to acquire data in almost real-time basis, and to process the data at the robot's local system. Therefore the end-users just need to connect their terminals to any network and then pointing the browser to the assigned address to display the analyzed results as graphs etc. While at same time they can control, also over web, the robot's movement, direction etc.

In this paper we focus on discussing our original concept on modular wireless robot, and followed with the main system of LWR which is responsible for control and monitoring the whole system. Detail discussion on the censors, real-time data processing, mechanical aspects and its integrated web interface will be presented elsewhere.

This paper is organized as follows : first of all we present our concept, and followed by the design of control and monitoring system. Finally we summarize the result and discuss future plans.

## 2. The concept

First of all, we have developed LWR to be as modular as possible to make it flexible to users' needs. This concept is depicted in Fig. 1, consisting of three modules : 1) main unit, 2) data-acquisition and 3) data processing modules. So, the system is divided into two independent units : the main unit and the unit of data acquisition and its processing software. Let us call the second unit briefly as DAPS throughout the paper. In principle, one might have single main unit with different packages of DAPS, or vice versa according to the needs. This characteristic would enable users to easily replace, for instance the censors and relevant data processing modules with any available packages later on with the same main unit.

Since we are going to open our main unit and its architecture freely, further this approach could encourage third parties to develop independently any relevant packages designed for some particular needs. Inversely, ones interested more in hardware developments, might build alternative main unit with different type of mechanics for robot, but with same architecture for the rest to keep its compatibilities with existing DAPS packages. Some considerable packages are, for example, censors of dangerous gas combined with software based chemical compound analyzer, vibration censor combined with software based seismograph, and so forth. This is the reason we call it as a generic robot.

We would like to emphasize that this concept is quite new and a breakthrough in robotic (hardware) technology that is actually inspired by the habit of open-source communities. Because, most of robots are usually constructed for single particular task.

According to the above concept, in order to realize high degree of freedom for both users and developers, and to keep full compatibilities in the future developments, let us list common characteristics should be fulfilled by LWR :

1. All aspects are fully controllable wirelessly over web through TCP/IP protocol.
2. Acquired data are stored and processed at the robot's local system independent from external apparatus.
3. The processed data can be retrieved and analyzed by users also over web such that no need for additional software installation at the user's terminal.
4. The hardware driven parts are replaced as much as possible with the software driven systems, even in the main unit.

Due to limited space, in the next section we are discussing only the microcontroller based hardware to enable data acquisition and remote mechanical control. Before going on let us mention some benefits in deploying this approach :

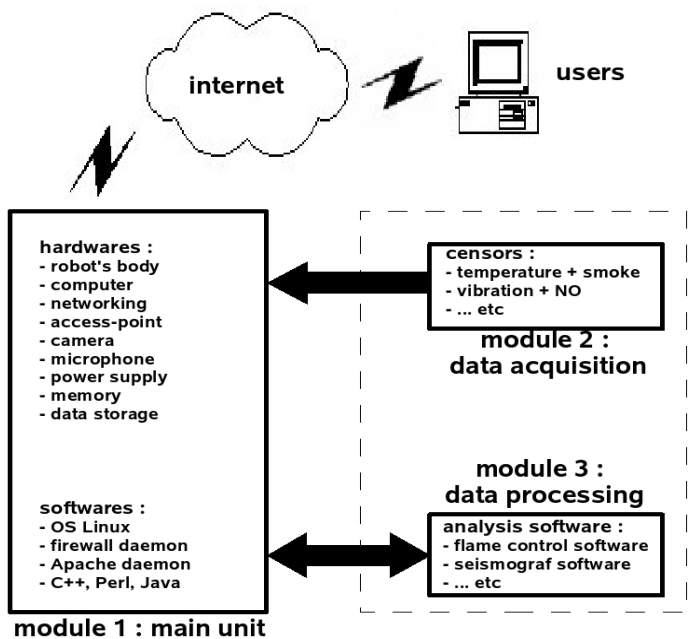

**Figure 1.** The block diagram of three modules in LWR.

- Easy access for end-users regardless the operating system used. There is also no need to install any additional software in terminal accessing the robot.
- Overall cost reduction, since most parts work on software-based system.
- High compatibility due to limited proprietary hardwares, and then embedded software, used in the system. Moreover all software-based parts are developed using freely available open-source softwares. In our case we use Debian Linux for the operating system, Apache for the web-server and some GNU Public License development languages.
- Highly safe at any untouchable areas of human being, since the robot is remotely controllable.

## 3. The data acquisition and control modules

As mentioned in the preceding section, we are focused only on the hardware aspects of data acquisition and control modules. Further discussion is based on the case of LWR as a prototype for the current approach.

Since there are two main tasks related to hardware in the system, that is control and data acquisition, we have deployed two microcontroller based devices. This can be seen in Fig. 2. Both are connected to the PC-based device, either a regular or mini computer over serial and parallel ports. Before moving on, we would like to comment on this choice.

Usually, most people tend to use the I/O port by applying a GPIB or data acquisition card due to its reliabilities and modularities. This option provides great benefit for users in simplifying the whole hardwares and in saving the design time as well, since there is no need to design everything from the scratch. Moreover one can further programme and interface it with the bundled softwares and its libraries for some popular low level languages as LabView or Visual Basic. All of these would enable creating attractive interfaces on the screen and store the acquired data with much less effort. On the other hand, this approach is lack of flexibility and freedom, especially for those who need to realize special tasks that cannot be accommodated by the standard libraries. Of course ones could develop their own libraries to solve this problem. However this requires more efforts that might be comparable to developing the hardwares from the scratch. Also, there is a risk of high dependency on certain products which could bind the users forever. Therefore, we decide to utilize alternative connection port and develop the hardwares from the scratch to meet specific needs of LWR and also to reduce the overall cost as well. We have chosen the parallel and serial ports. Using those ports, there is no need to design nor purchase expensive I/O port. We just need to plug the devices we have developed into the ports and programme them to be anything matches our purposes. Importantly, this can be used not only for control, but also for data acquisition depending on the programme.

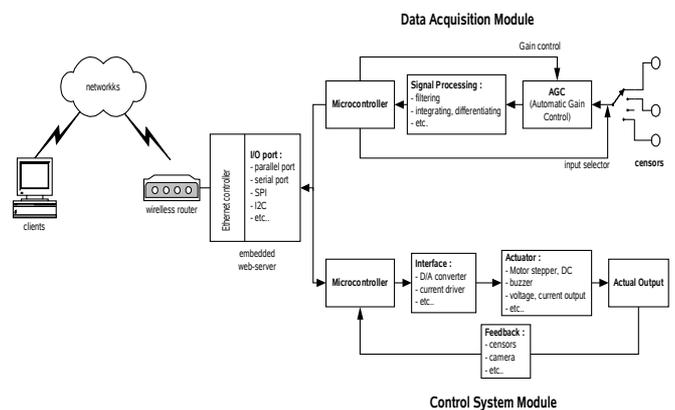

**Figure 2.** The block diagrams for data acquisition and control.

In principle both devices can be integrated in a single microcontroller. However, to reduce the delays due to the limitation of microcontroller, we divide them into two independent devices. Another reason is for the sake of modularity as mentioned before. Because the censors embedded in data acquisition module, while the control system is integrated in main unit together with the embedded computer. We should emphasize that the control system also involves some monitoring tasks like feedbacks from the wheels to calculate the actual distance, etc. Therefore we use serial port which has single incoming and outgoing data communication for the control system, and parallel port which have multi channels for the data acquisition. Because one in general may attach multi censors for a comprehensive observation.

Now let us discuss the way to control the movement of LWR. Instead of using the images captured by camera which would require complicated and resource wasting image processing, we combine together the compass censor and its wheels to recognize the current position. The compass censor provides information of the actual angle against the North-South pole. While calculating the rotation of the wheels we can obtain the point-to-point distance. This mechanism is integrated on the web as virtual compass and the footprint of wheels from one point to another. Of course we also attach a small camera on the robot, but only for recognizing the actual visualization around the robot.

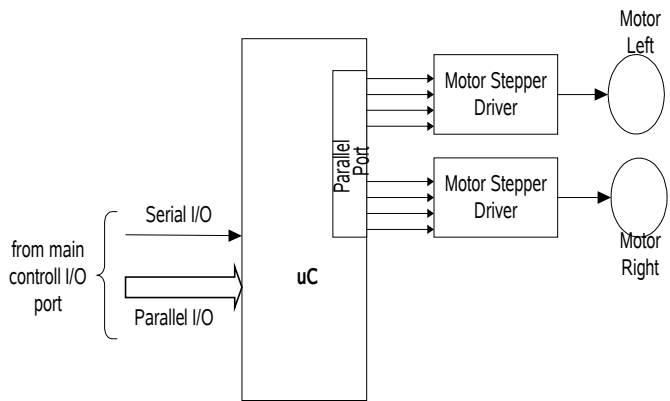

**Figure 3.** Control module.

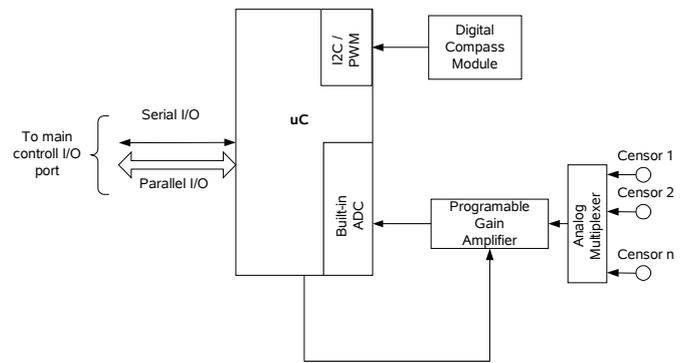

**Figure 4.** Data acquisition module.

The block diagram for control module is given in Fig.. 3 in the simplest case of mechanical robot with regular wheels. This module is responsible only to control the actuator. In this design we set the motor stepper as an actuator. As an illustration let us concern only two motors as wheels. By using these motors we are able to control the robot movement in four directions such as turning left and right, moving forward and backward. However the electricity current supplied by microcontroller is too small to accelerate the motor steppers directly. In order to overcome this problem, we must use motor driver to amplify the current. One of the most popular chip suitable for this purpose is ULN2803. This IC has eight darlington array to amplify the current. By using only one parallel port in microcontroller such as PA,PB,PC or PD, we can control two motors steppers simultaneously [5,6]. The following code is used to send a command to microcontroller to move the robot in a particular direction :

```
#include <xxx.h>        //it depend on microcontroller used
void main()
{
//input port for capturing the command sent by parallel port
//can be set depend on capabilities of each microcontroller
    while(1)            //loop forever
    {
    data=input();       //read command into variable data
    switch(data)
        {
        case '1' : turn left;
        case '2' : turn right;
        case '3' : moving forward;
        case '4' : moving backward;
        }}
}
```

In contrast, the data acquisition module is used to retrieve some physical parameters from attached multiple censors for temperature, gases and so forth. As discussed previously, there is also another censor related to the control system, that is the digital compass to acquire the current angle against North-South pole. Therefore, we have designed to divide the data acquisition module into two parts as shown in Fig. 4. The first one involves some censors with digital outputs like compass censor. The other is responsible for censors producing analog outputs such as temperature and gases censors.

For the compass censor we have deployed the digital compass module CMPS03. This module produces serial digital output data that can be read by microcontroller either in PWM or I2C modes [7]. In the sense of modularity, it might be better to split this part off the data acquisition module in real application. Because the compass censor should belong to the main unit to control the robot itself.

The temperature and gases censor are processed by microcontroller using internal ADC as argued previously. In this design we use microcontroller Atmega8535 from Atmel. Due to different level of each output of censors, the programmable gain amplifier is used. The gain is controlled by microcontroller to distinguish from which censor the data is taken. It is clear that simultaneous data retrieval from multiple censors is impossible. However, by adjusting the sequence of data retrieval from each censor, we may obtain data in an appropriate interval for each of them. Although this is not a real-time process, concerning the actual required time interval, that is typically much longer around a minute, we can say that the current system has already realized a "real-time" data acquisition.

We would like to stress that all microcontroller based-hardwares above are utilized moreless only for converting the analog data to be the digital ones. All data filtering are handled by software embedded in data processing, and then DAPS, module. This point is important since there might be multiple censors attached in DAPS which require different signal processings. Again, data processing module carries out not only the task of filtering the incoming signals, but also analyzing the whole data to be understandable results for users as graphs, tables etc.

**4. Summary**

We have developed the LIPI Wireless Robot as a typical prototype for newly introduced concept of modular wireless robot. The robot has unique characteristics adjustable for various purposes. We argue that LWR is quite efficient and has good total cost-performance due to its modularity and dominant software based solutions. The system has complicated aspects require further developments, and will be presented elsewhere like :
- Web-based real-time control for robot's movement using compass censor and wheel rotation.
- Web-based software to analyze the acquired data from particular DAPS packages in real-time basis.

- A new or modified algorithms for software-based control on robot's mechanics.

In the future we also plan to develop another main unit and its robot. Especially to improve mechanical aspects of LWR to enable more advanced and smooth movements. Rather than hardware-based approach, this will be done by utilizing integrated and software-based algorithms.

Lastly, we would like to announce that the architecture and all related softwares of LWR will be open for public under GNU Public License once we consider the system is ready for further development by open-source communities around the world.

**Acknowledgment**

This work is partially supported by the Riset Kompetitif LIPI in fiscal year 2007 under Contract no. 11.04/SK/KPPI/II/2007.

**References**

[1] A. Ferworn, R. Roque, and I. Vecchia, "MAX: Teleoperated Dog on the World Wide Web", Proc. of the 2nd International Workshop on Presence, Colchester, UK, 1999.

[2] A. Ferworn, R. Roque, and I. Vecchia, "MAX: Wireless Teleoperation via the World Wide Web", Proc. of the 1999 IEEE Canadian Conference on Electrical and Computer Engineering, Edmonton, Alberta, Canada, 1999.

[3] WAX's webpage, http://ncart.scs.ryerson.ca/wax.

[4] LIPI Wireless Robot, *http://robot.teori.fisika.lipi.go.id*.

[5] Thomas Braunl, "*Embedded Robotic, Mobile Robot Design and Applications with Embedded System*", Springer, 2003.

[6] Gordon McComb, Myke Predko, "*Robot's Builder Bonanza*", McGraw-Hill, 2006

[7] Jan Axelson, "*The Microcontroller Idea Book*", Lakeview Research, 1997.